%% file: main.tex
\documentclass{article}

\usepackage[preprint]{neurips_2025}

\usepackage[utf8]{inputenc} 
\usepackage[T1]{fontenc}    
\usepackage{hyperref}       
\usepackage{url}            
\usepackage{booktabs}       
\usepackage{amsfonts}       
\usepackage{nicefrac}       
\usepackage{microtype}      
\usepackage{xcolor}         
\usepackage{algpseudocode}
\usepackage{graphicx}
\usepackage{algorithm} 
\usepackage{amsmath}
\usepackage{tablefootnote}
\usepackage{subcaption}
\usepackage{arydshln}

\input{preamble.tex}
\newcommand{\black}[1]{\textbf{\textcolor{black}{#1}}}
\newcommand{\bred}[1]{\textbf{\textcolor{red}{#1}}}

\title{STR-Match: Matching SpatioTemporal Relevance \\ Score for Training-Free Video Editing}

\author{%
  Junsung Lee$^1$\hspace{1cm}Junoh Kang$^1$\hspace{1cm}Bohyung Han$^{1,2}$\\
  ECE$^1$ \& IPAI$^2$, Seoul National University\\
  \texttt{\{leejs0525, junoh.kang, bhhan\}@snu.ac.kr} \\
}

\begin{document}

\maketitle

\input{sections/abstract}
\input{sections/introduction}

\input{sections/related_work}
\input{sections/preliminary}
\input{sections/method}
\input{sections/experiment}
\input{sections/conclusion}


\bibliography{main}
\bibliographystyle{unsrt}

\newpage
\appendix
\input{sections/supple_.tex}

\end{document}

%% file: preamble.tex
\usepackage{multirow}
\usepackage{xspace}
\usepackage{caption}
\usepackage{thm-restate}
\usepackage{footmisc}
\usepackage[capitalize,noabbrev]{cleveref}
\usepackage{wrapfig}
\usepackage{enumitem}

\newcommand{\eg}{\textit{e}.\textit{g}.\xspace}

\newcommand{\x}{\mathbf{x}}
\newcommand{\z}{\mathbf{z}}

\newcommand\myldots{\ifmmode\ldots\else\makebox[0.5em][c]{.\hfil.\hfil.}\thinspace\fi}
\newcommand\mycdots{\ifmmode\cdots\else\makebox[0.5em][c]{.\hfil.\hfil.}\thinspace\fi}

%% file: sections/abstract.tex

\begin{center}
	\includegraphics[width=\textwidth]{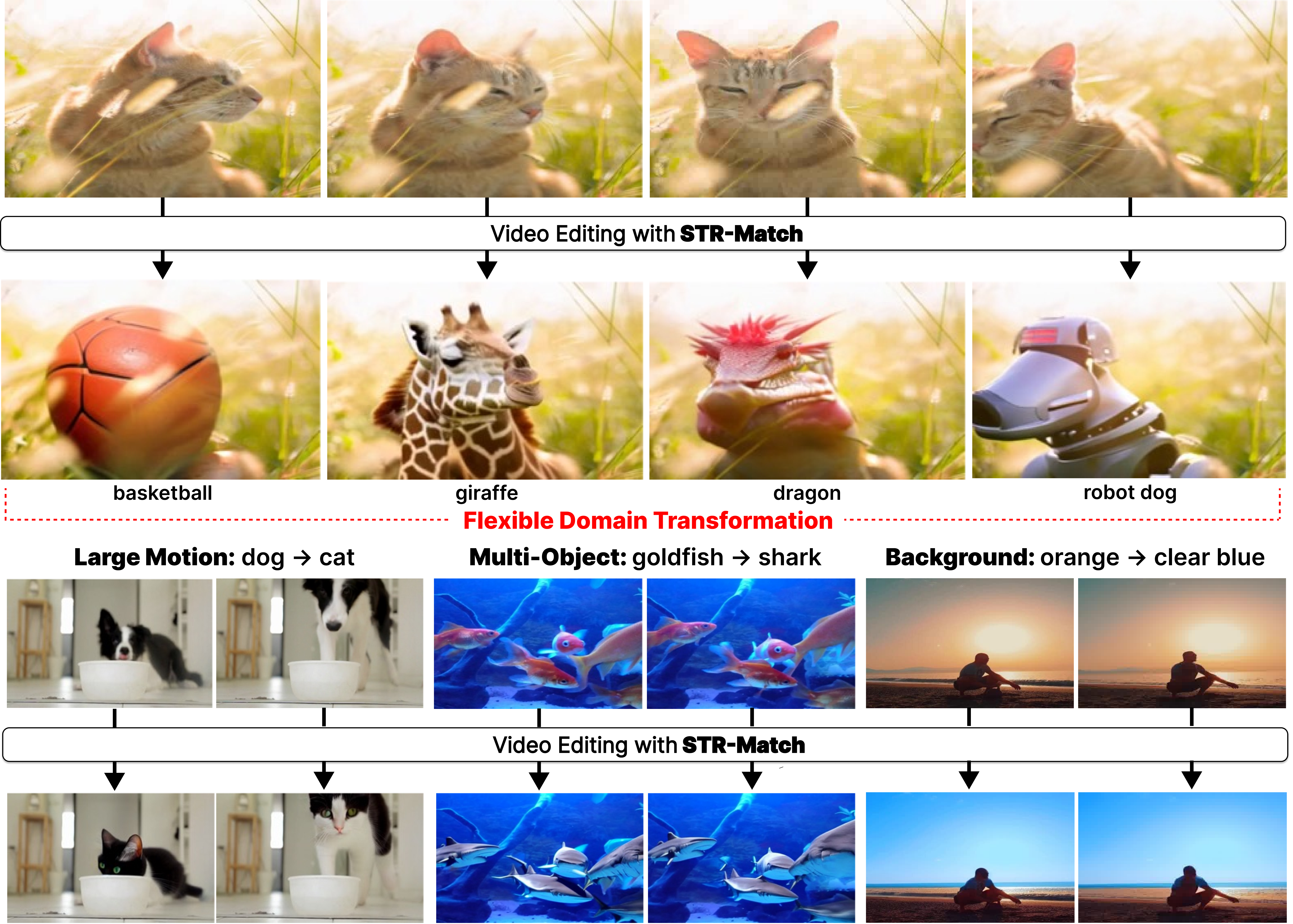}
	\captionof{figure}{
	\textbf{Generated videos using our proposed algorithm, STR-Match.}
	Our proposed algorithm, STR-Match, successfully performs flexible domain transformations while preserving the visual information of the source video during the video editing process.
	It is also applicable to various scenarios, including large motion, multi-object, and background editing.
	}
	\label{fig:thumbnail}
\end{center}

\begin{abstract}

Previous text-guided video editing methods often suffer from temporal inconsistency, motion distortion, and—most notably—limited domain transformation.
We attribute these limitations to insufficient modeling of spatiotemporal pixel relevance during the editing process.
To address this, we propose STR-Match, a training-free video editing algorithm that produces visually appealing and spatiotemporally coherent videos through latent optimization guided by our novel STR score.
The score captures spatiotemporal pixel relevance across adjacent frames by leveraging 2D spatial attention and 1D temporal modules in text-to-video~(T2V) diffusion models, without the overhead of computationally expensive 3D attention mechanisms.
Integrated into a latent optimization framework with a latent mask, STR-Match generates temporally consistent and visually faithful videos, maintaining strong performance even under significant domain transformations while preserving key visual attributes of the source.
Extensive experiments demonstrate that STR-Match consistently outperforms existing methods in both visual quality and spatiotemporal consistency.

Project page: \url{https://jslee525.github.io/str-match}

\end{abstract}

%% file: sections/introduction.tex

\vspace{-2mm}
\section{Introduction}

Diffusion models~\cite{ddpm, ddim, scorebased} are the leading framework for high-fidelity image and video generation using text prompts. 
Their applications now extend to tasks such as text-guided image and video editing, where the goal is to generate outputs aligned with target text prompts while preserving regions consistent with both the source and target prompts in the original content.
The overall process of text-guided image editing typically involves generating a target image guided by information extracted during the forward or reconstruction process of the source image---most often through latent optimization or attention injection, though a few methods adopt alternative approaches~\cite{masactrl, p2p, pnp, pix2pixzero, hslee, contrastive, imagic, pic}.

While text-guided image editing methods have demonstrated impressive editing capabilities, directly applying them to video editing presents several challenges, including frame inconsistency and undesired motion change.
To achieve strong video editing performance while addressing these issues, many prior works~\cite{fatezero, gav, flatten, videograin} leverage pretrained text-to-image (T2I) models augmented with additional components.
Some other recent works~\cite{motionflow, dmt, motionconsistency, uniedit} adopt text-to-video (T2V) models to tackle these problems.
However, these methods still suffer from the same issues and exhibit degraded performance in challenging scenarios (\textit{e.g.,} large domain shifts). 


These limitations in text-guided video editing stem from inadequate modeling of spatiotemporal pixel relevance, which is crucial for producing natural and coherent video content.
To address these challenges, we introduce STR-Match, a training-free algorithm that generates videos via latent optimization guided by a novel STR score. 
The STR score, defined as the multiplicative combination of self- and temporal-attention maps, captures spatiotemporal pixel relevance across adjacent frames by combining 2D spatial and 1D temporal attention from a text-to-video (T2V) diffusion model, without relying on costly 3D attention mechanisms. 
This joint formulation enables more effective optimization than using the attention components separately, ultimately improving video quality.
Integrated into a latent optimization framework with a masking strategy, STR-Match produces temporally consistent, high-fidelity outputs, effectively handling challenging editing cases and maintaining the key visual attributes of the source.

Our primary contributions are summarized as follows:
\begin{itemize}[leftmargin=0.8cm]

    \item We introduce STR-Match, a novel training-free text-guided video editing approach built upon pretrained T2V diffusion models. It matches spatiotemporal information in the generation process (target latents) to that of the forward process (source latents) via latent optimization, optionally incorporating a latent masking strategy for improved preservation of source content.
    This design addresses key limitations of existing methods stemming from insufficient modeling of spatiotemporal pixel relevances.

    \item 
    To obtain spatiotemporal information, we propose the STR score, a spatiotemporal pixel relevance score that combines self- and temporal-attention maps without requiring full 3D attention.
    The STR score also enables flexible optimization, resulting in enhanced overall video quality.
    

    
    \item Through extensive experiments on various video editing tasks, we demonstrate that STR-Match outperforms existing training-free video editing approaches both quantitatively and qualitatively. 
    STR-Match generates temporally coherent, high-fidelity videos with flexible domain transformations, while preserving the visual integrity of the source video. 
    It consistently outperforms prior methods in these aspects.
    


    
    
    

	
\end{itemize}

%% file: sections/related_work.tex
\section{Related works}

\subsection{Text-to-video diffusion model}
\label{subsec: t2v}

Recent works~\cite{videocrafter2, Lavie} build on diffusion models by extending pretrained text-to-image (T2I) architectures.
These methods commonly introduce lightweight 1D temporal modules into 2D spatial backbones, enabling efficient video generation while preserving the visual priors learned from T2I models.
While previous T2V models such as VideoCrafter2~\cite{videocrafter2} and LaVie~\cite{Lavie} extend pretrained T2I architectures by inserting lightweight temporal modules into 2D spatial backbones, more recent approaches aim to capture richer spatiotemporal pixel relevances through full 3D attention.
Building on advances in efficient attention computation frameworks such as xFormers~\cite{xformers} and FlashAttention~\cite{flashattn}, the latest T2V models~\cite{cogvideox, opensora} incorporate 3D full attention into their architectures.
For example, CogVideoX~\cite{cogvideox} and Open-Sora-2.0~\cite{opensora} adopt 3D autoencoding architectures with integrated 3D full attention, leveraging FlashAttention to enable efficient attention computation.
However, these models typically compute attention outputs without explicitly retaining attention maps, which limits their applicability in tasks requiring controllable attention—such as fine-grained video editing.


\subsection{Training-free video editing methods}
\label{subsec: priorwork}

\paragraph{T2I-based video editing methods} 
With the rapid progress of image editing works~\cite{masactrl, p2p, pnp, pix2pixzero, hslee, contrastive, imagic, pic}, recent works~\cite{fatezero, gav, flatten, videograin} leverage pretrained T2I models with addtional components to complement frame consistency.
FateZero~\cite{fatezero} manipulates attention maps using binary masks from cross-attention and improves temporal consistency by warping middle-frame features during diffusion.
Ground-A-Video~\cite{gav} leverages external models—such as GLIGEN~\cite{gligen}, RAFT~\cite{raft}, ZoeDepth~\cite{zoedepth}, and ControlNet~\cite{controlnet}—to guide attention modulation with attention maps.
FLATTEN~\cite{flatten} manipulates attention maps to follow patch trajectories derived from optical flow~\cite{raft}, aiming to maintain frame consistency.
VideoGrain~\cite{videograin} modulates both self- and cross-attention to address multi-grain video editing tasks, relying on external methods~\cite{pnp, flatten} to enhance frame consistency.
Although these T2I-based methods have demonstrated strong editing capabilities, they still struggle from temporal inconsistency and motion distortion.
Moreover, many of these approaches rely on attention injection, which can disrupt the computational graph of the pretrained model and often lead to visual artifacts. 


\vspace{-2mm}
\paragraph{T2V-based video editing methods} 
In contrast to T2I-based approaches, several recent methods~\cite{motionflow, dmt,  motionconsistency, uniedit} leverage pretrained T2V models to address temporal consistency in the video editing task.
For example, DMT~\cite{dmt} utilizes a pretrained T2V model and introduces a feature descriptor extracted from intermediate layers to guide latent optimization for motion preservation.
MotionFlow~\cite{motionflow} incorporates losses from cross-, self-, and temporal-attention, along with mask-based manipulation, to preserve motion information in the source video.
Zhang et al.~\cite{motionconsistency} extracts motion patterns using temporal modules and applies a frame-to-frame consistency loss during generation. 
These approaches utilize latent optimization, which preserves the pretrained model's computational process, allowing for smoother outputs with fewer visual artifacts.
However, these methods primarily focus only on motion guidance, which often leads to modifications in unwanted regions (\textit{e.g.}, backgrounds).
While UniEdit~\cite{uniedit} attempts to address these issues by applying attention injection to edit appearance or motion in source videos, it often suffers from texture misalignment in the foreground and background regions.


%% file: sections/preliminary.tex

\section{Preliminary}
\label{sec:preliminary}

\textbf{Text-to-video diffusion model} 
We summarize the basic concept of pretrained text-to-video diffusion models as we use the models to perform text-guided video editing. 
The key components of text-to-video model is an encoder $\text{Enc}(\cdot)$, a decoder $\text{Dec}(\cdot)$, and a noise prediction network $\epsilon_\theta(\cdot)$.
Encoder spatially and temporally compresses a video vector $\x \in \mathbb{R}^{F \times H \times W \times 3}$ to a latent vector $\z_0 \in \mathbb{R}^{f \times h \times w \times c}$, and decoder decompresses the latent vector to the video vector.
The noise prediction network learns the distribution of latent vectors $\z_0$, and is trained to minimize following objective function:
\begin{align} \label{eq:diffusion_objective}
	\mathbb{E}_{\z_0,\mathbf{c},t,\epsilon} [||\epsilon_\theta(\z_t,t,\mathbf{c}) - \epsilon||_2^2],
\end{align}
where $\z_0$ denotes the video latent, $\mathbf{c}$ is the corresponding text prompt, $t$ is the diffusion timestep, and $\z_t = \alpha_t \z_0 + \sigma_t \epsilon$ for $\epsilon \sim \mathcal{N}(\mathrm{0}, \mathrm{I})$. 
$\alpha_t$ and $\sigma_t$ are predifined constants satisfying $\alpha_0 = 1$, $\sigma_0=0$, and $\sigma_T/\alpha_T \gg 1$.

\textbf{Attention modules} 
While text-to-video diffusion models extend the success of text-to-image models by incorporating temporal modules, we specifically focus on two critical features: spatial self-attention map and temporal-attention map. 
Spatial self-attention map, whose dimension is $\mathbb{R}^{f \times h \times n \times n}$, captures relevances between pixels within each frame, where $f$ denotes the number of frames, $n$ represents the number of pixels per frame, and $h$ indicates the number of attention heads. 
For the rest of the paper, we denote $p,q \in \{1,2,...n\}$ for the spatial location of pixel and $i,j \in \{1,2,...f\}$ for the frame number. 
Combining these, $I_i(p)$ represents the pixel at location $p$ in $i$-th frame.
Then, the self-attention map element $\text{Attn}(I_i(p)\rightarrow I_i(q))$ can be interpreted as importance of $I_i(q)$ to $I_i(p)$ in 2D spatial space. 
Similarly, temporal-attention map, whose dimension is $\mathbb{R}^{n \times h\times f \times f }$, encodes inter-frame relevances for each pixel, and the element $\text{Attn}(I_i(p)\rightarrow I_j(p))$ represents the importance of $I_j(p)$ to $I_i(p)$ in 1D temporal space.

%% file: sections/method.tex


\section{Methods}
\label{sec:method}

\begin{figure}[t]
	\centering
	\includegraphics[width=\linewidth]{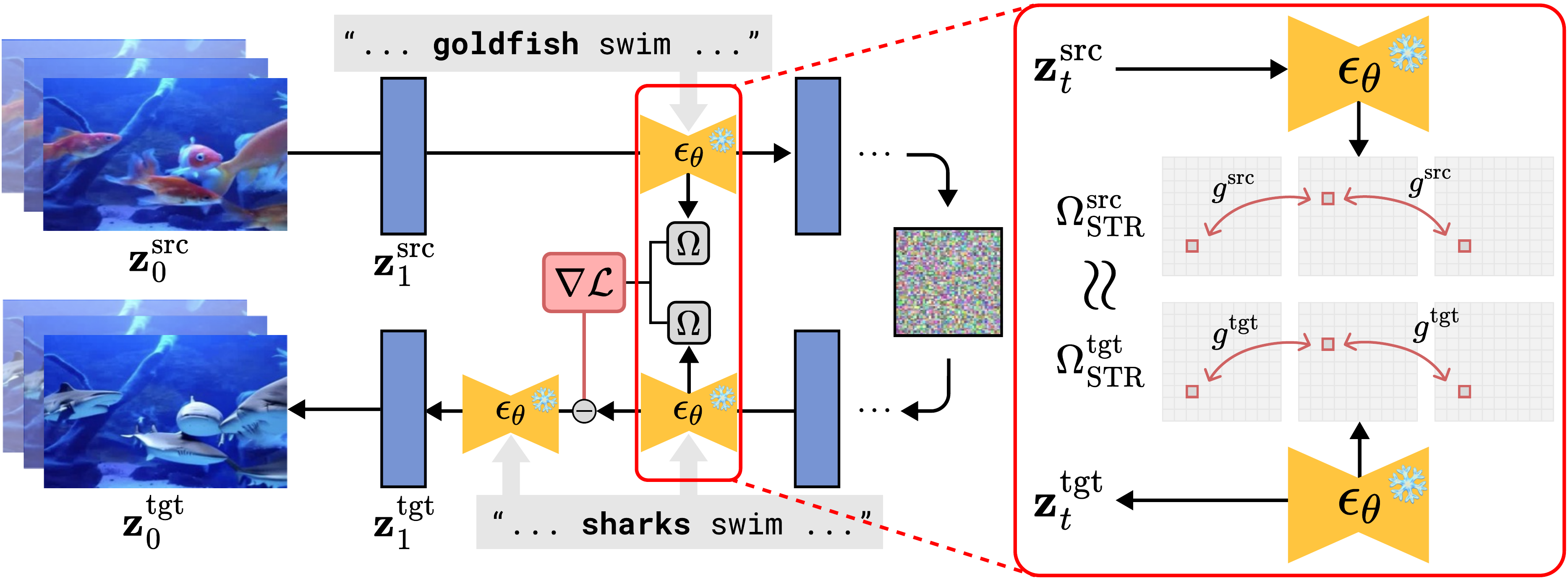}
	\caption{
	\textbf{Illustration of overall STR-Match framework.} 
	We first perform a forward diffusion process, and extract the STR score $\Omega^{\text{src}}_{\text{STR}, t}$ from the source video.
	Then, the target latent is initialized as $\mathbf{z}^{\text{tgt}}_T = \mathbf{z}^{\text{src}}_T$, and during the generation process, we extract the target STR score $\Omega^{\text{tgt}}_{\text{STR}, t}$ and optimize the latent $\mathbf{z}^{\text{tgt}}_t$ using a negative cosine similarity between the source and target STR scores.
	To further preserve unediting regions, we optionally apply a latent mask strategy using a binary mask $M$.
	\vspace{-5mm}
	}
	\label{fig:stfigure}
\end{figure}

Many text-guided image editing methods~\cite{masactrl, p2p, pnp, pix2pixzero, contrastive} manipulate attention maps, demonstrating that modeling pixel relevances is crucial for effective image editing.
Likewise, we expect that spatiotemporal pixel relevances in videos are essential for effective video editing. 
To this end, we propose the STR score, which captures spatiotemporal relevances between pixels across different frames by leveraging self- and temporal-attention maps.
It is an aggregation of bidirectional pixel relevances across adjacent frames, efficiently capturing spatiotemporal information and enabling the extraction of key visual attributes from the source video.
By integrating the STR score into a latent optimization framework, as illustrated in \cref{fig:stfigure}, we enable video editing that preserves source content while achieving high visual quality with flexible domain shifts.

\subsection{STR score: SpatioTemporal Relevance score}
\label{subsec:st-feature}



\begin{figure}[t]
	\centering
	\includegraphics[trim=0cm 0cm 0cm 0.3cm, clip, width=\linewidth]{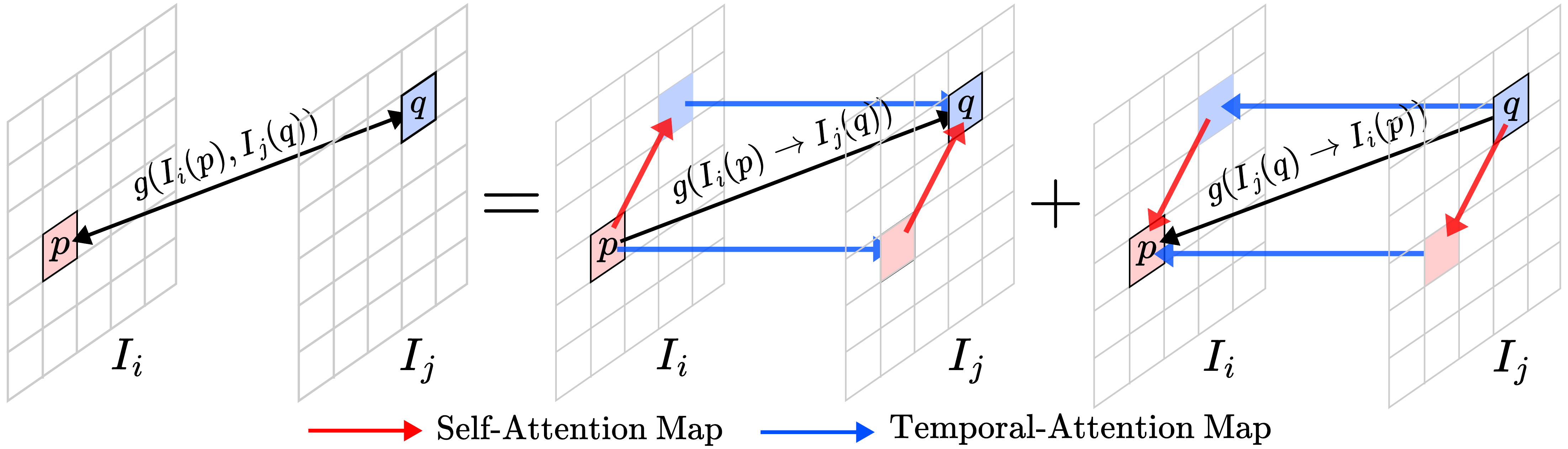}
	\caption{
		\textbf{Illustration of STR score.}
		(Left) The bidirectional pixel relevance in the spatiotemporal space $g(I_i(p), I_j(q))$ is computed by summing two directional relevance scores along opposite directions.
		(Right) Each figure illustrates the directional pixel relevance, $g(I_i(p) \rightarrow I_j(q))$ and $g(I_j(q) \rightarrow I_i(p))$, both of which are computed solely through pixel-wise multiplication of self- and temporal-attention maps.
	}
	\label{fig:approx_joint}
\end{figure}

To quantitavely represent relevance between two pixels $I_i(p)$ and $I_j(q)$ in spatiotemporal space, we define two functions: bidirectional relevance $g(\cdot, \cdot)$, and directional relevance $g(\cdot \rightarrow \cdot)$.
The directional relevance $g(I_i(p) \rightarrow I_j(q))$ quantifies the importance of $I_j(q)$ to $I_i(p)$ in spatiotemporal space, and intuitively, it is expected to be large if both the importance of $I_j(p)$ to $I_i(p)$ and $I_j(q)$ to $I_j(p)$ are high, or the importance of $I_i(q)$ to $I_i(p)$ and $I_j(q)$ to $I_i(q)$ are high.
From this motivation, we define directional relevance between $I_j(q)$ given $I_i(p)$ as 
\begin{align}
    g(I_i(p) \rightarrow I_j(q)) 
    &:= \text{Attn}(I_i(p) \rightarrow I_j(p))\, \text{Attn}(I_j(p)\rightarrow I_j(q)) \nonumber \\
	&~~ + \text{Attn}(I_i(p)\rightarrow I_i(q))\, \text{Attn}( I_i(q) \rightarrow I_j(q)),
    \label{eq:rel_cond2}
\end{align}
for $\text{Attn}(\cdot \rightarrow \cdot)$ defined in \cref{sec:preliminary}.
The bidirectional relevance $g(I_i(p), I_j(q))$ extends the directional relevance by considering the connection between $I_i(p)$ and $I_j(q)$ in both directions, as illustrated in \cref{fig:approx_joint}.
Specifically, it is defined as a sum of the importance of $I_j(q)$ to $I_i(p)$ and the importance of $I_i(p)$ to $I_j(q)$:
\begin{align}
	g(I_i(p), I_j(q)) := g(I_i(p) \rightarrow I_j(q)) + g(I_j(q) \rightarrow I_i(p)). \label{eq:def_3d}
\end{align}
Notably, the bidirectional relevance is fully computed from self- and temporal-attention maps without requiring any additional training or models.

To capture spatiotemporal information in the source video—such as motion and structural layout—we aggregate bidirectional pixel relevances across adjacent frames into a unified representation, termed the STR score.
The STR score $\Omega_{\text{STR}}$, or spatiotemporal pixel relevance, is formally defined as follows:
\begin{equation}
	\label{stwindow}
	\Omega_{\text{STR}}(i, p, q)= \sum_{j\in \mathcal{N}(i)} g(I_i(p), I_j(q)), 
\end{equation}
where $\mathcal{N}(i)$ is a set of neighboring frame numbers to the $i$-th frame. 


\subsection{Overall framework: STR-Match}
\label{subsec:overall}


The overall procedure of our method is illustrated in \cref{fig:stfigure} and \cref{alg:stflow}. 
We first solve the forward diffusion process of the source video.
During the forward process, we extract STR scores $\Omega^{\text{src}}_{\text{STR}, t}$ at every timestep and noisy latent $\z_T^{\text{src}}$.
Then, starting from $\z_T^{\text{tgt}} = \z_T^{\text{src}}$ as initial point, we perform generation process with latent optimization.
For each denoising step, we first optimize the latent variable $\z_t^{\text{tgt}}$, and then solve diffusion process with the optimized latents.
The optimization is performed with the following equation:
\begin{equation}
	\z^\text{tgt}_t \leftarrow \z^\text{tgt}_t - \lambda\nabla_{\z^\text{tgt}_t}\mathcal{L}_{cos}(\Omega^\text{src}_{\text{STR}, t}, \Omega^\text{tgt}_{\text{STR}, t}),
	\label{eq:opt}
\end{equation}
where $\mathcal{L}_{cos}$ is a negative cosine similarity, and $\lambda$ is a hyperparameter for controlling the guidance strength.
The equation is designed to maximize the cosine similarity between the source and target STR scores, encouraging the spatiotemporal pixel relevances in the target video to align with those of source video to promote preservation of spatiotemporal information. 

Since the optimization process preserves the computational graph of the pretrained model, it enables the generation of smooth, high-quality videos while maintaining key visual information from the source.
Moreover, since $\Omega_{\text{STR}}$ is conceptually defined as the element-wise product of self- and temporal-attention maps, it enables more flexible optimization compared to using them independently, thereby further enhancing video quality.

\vspace{-2mm}
\paragraph{Latent mask strategy} 
To better preserve regions that are not intended to be edited (\textit{e.g.}, backgrounds), we mix the optimized latent with the latent obtained during the forward process at the same timestep.
For a binary mask $M$, where values are $1$ for regions to be edited and $0$ otherwise, and latents $\z_{t}^{\text{src}}$ obtained during the forward diffusion process of source video, the final target latents are updated as 
\begin{align}
	\z_t^{\text{tgt}} \leftarrow (1-\texttt{dilate}(M)) \odot \z_t^{\text{src}} + \texttt{dilate}(M) \odot \z_t^{\text{tgt}}.
	\label{eq:mask}
\end{align}
This masking strategy ensures to preserve non-target regions in the source video during editing.
The latent binary mask is resized and dilated version of segmentation map of editing region of the source video.
The \texttt{dilate} function is applied to help flexible shape modification. 


\begin{algorithm}[t]
	\caption{STR-Match}\label{alg:stflow}
	\begin{algorithmic}[1]
		\State \textbf{Input}: $\z^{\text{src}}_0$ (source video), $p^{\text{src}}$ (source prompt embedding), $p^{\text{tgt}}$ (target prompt embedding), $\Phi(\cdot)$ (ODE solver), $M$ (foreground binary mask, optional)
		\State \textbf{Hyperparameter}: $\lambda$ (coefficient of negative cosine similarity)
		
		\For{$t=0$ \textbf{to} $T-1$} 
		\State $\epsilon^\text{src}_t \gets \epsilon_\theta(\z^{\text{src}}_t, t, p^{\text{src}})$
		\State Compute and save $\Omega^{\text{src}}_{\text{STR}, t}$ from $\epsilon_\theta(\cdot)$
		\State $\z^{\text{src}}_{t+1} \gets \Phi(\z^{\text{src}}_t, \epsilon^{\text{src}}_t, t\rightarrow t+1)$ 
		\EndFor
		
		\State $\z_T^{\text{tgt}} \gets \z_T^{\text{src}}$
		
		\For{$t=T$ \textbf{to} $1$}
		\State {Obtain} $ \epsilon_\theta(\z^{\text{tgt}}_t, t, [p^{\text{tgt}}; p^{\text{src}}])$
		\State {Compute} $\Omega^{\text{tgt}}_{\text{STR}, t}$ { from} $\epsilon_\theta(\cdot)$
		\State $\z^{\text{tgt}}_t \gets \z^{\text{tgt}}_t - \lambda\nabla_{\z^{\text{tgt}}_t} \mathcal{L}_{cos}(\Omega^{\text{src}}_{\text{STR}, t}, \Omega^{\text{tgt}}_{\text{STR}, t})$
		\State $\epsilon^{\text{tgt}}_t \gets \epsilon_\theta(\z^{\text{tgt}}_t, t, [p^{\text{tgt}};p^{\text{src}}])$
		\State $\z^{\text{tgt}}_{t-1} \gets \Phi(\z^{\text{tgt}}_t, \epsilon^{\text{tgt}}_t, t\rightarrow t-1)$ 
		\If{\textit{use latent mask}}
		\State $\z^{\text{tgt}}_{t-1} \gets (1-\texttt{dilate}(M)) \odot  \z^{\text{src}}_{t-1}  + \texttt{dilate}(M) \odot \z^{\text{tgt}}_{t-1}$ 
		\EndIf
		\EndFor
		
		\State \textbf{Result}: $\z^{\text{tgt}}_0$ (target video)
	\end{algorithmic}
\end{algorithm}

%% file: sections/experiment.tex

\section{Experiments}
\label{sec:experiments}
\subsection{Implementation details}
\label{subsec:implementation}

Throughout the experiments, STR-Match is implemented using LaVie~\cite{Lavie} as the pretrained T2V model, with the hyperparameter $\lambda=0.01$, and optimized using SGD.
For extreme cases of qualitative results~(\textit{e.g.}, cat $\rightarrow$ basketball), we select $\lambda$ from the range $[0.005, 0.015]$.
For efficient inference, we extract the STR score based on self- and temporal-attention maps, excluding those from the finest resolution.
We compare our method against recent training-free video editing algorithms: FateZero~\cite{fatezero}, Ground-A-Video~(GAV)~\cite{gav}, FLATTEN~\cite{flatten}, VideoGrain~\cite{videograin}, DMT~\cite{dmt}, and UniEdit~\cite{uniedit}.
For T2I-based methods~(FateZero, Ground-A-Video, FLATTEN, VideoGrain), we follow their official implementations. 
For T2V-based baselines~(DMT, UniEdit), we adopt LaVie~\cite{Lavie} as the pretrained T2V model to ensure a fair comparison.
We employ a video segmentation model SAM-Track~\cite{samtrack} to obtain binary mask $M$, and OWL-ViT~\cite{bbox}, an object detection model to obtain bounding boxes for Ground-A-Video.
For more detailed description for base model and external model used in implementation, please see \cref{tab:details} in \cref{app:quant}.
The number of diffusion timesteps is set to $50$ and classifier-free guidance scale~\cite{classifier} is $7.5$.
We use L2 loss with $\lambda = 0.08$ for optimization using a concatenation of self- and temporal-attention maps in the ablation study, which we find to be an effective weight. 
For all experiments, we utilize a single NVIDIA L40S GPU with 48 GB of memory.

\input{tables/qual.tex}

\paragraph{Quantitative evaluation protocol}
For quantitative evaluation, we collect a total of 54 videos, each consisting of 16 frames, comprising samples from the TGVE dataset~\cite{tgve} and additional videos sourced from the Internet~\footnote{https://www.pexels.com}.
We utilize VideoLLaMA3-7B~\cite{videollama}, a pretrained video captioning model, to obtain concise prompts of source videos automatically, and randomly change nouns to construct the corresponding target prompt.
We measure four metrics to evaluate the fidelity and fatihfulness of the edited videos to source video and target prompt.
Frame Consistency~(FC) suggested in VBench~\cite{vbench} measures the smoothness of videos, leveraging motion priors in the frame interpolation model~\cite{frameinterpolation}. 
CLIP Similarity~(CS) computes the average CLIP score~\cite{clipscore} between the target prompt and edited video.
BG-LPIPS~(BL) calculates the Learned Perceptual Image Patch Similarity~(LPIPS) score~\cite{lpips} between maksed frames of the source and generated videos, where the mask is $1$ for regions to preserve.
Motion Error~(ME) quantifies the average motion difference between the source and generated videos.
It is calculated as the pixel-wise differences of optical flows between each video pair, where the optical flows are obtained using RAFT-Large~\cite{raft}.



\subsection{Qualitative results}
\label{subsec:exp_qual}

\paragraph{Key features of STR-Match}
%
\cref{fig:thumbnail} demonstrates STR-Match’s robust editing performance, highlighting its flexibility in challenging scenarios—such as transforming objects into entirely different categories, handling large motion, performing multi-object editing, and modifying background.
For instance, transforming a cat into a basketball or a giraffe demonstrates STR-Match’s ability to faithfully adapt object shapes to target prompts without being overly anchored to the original shape.
Moreover, changing a cat into a dragon or a robot dog—objects unlikely to appear in the original scene—illustrates STR-Match’s effective integration of edited elements with the background.
These examples emphasize how STR-Match manages domain-shifted objects and significant shape changes, while ensuring the edited elements blend naturally with the background. 
This combination of flexibility, visual quality, and motion preservation makes STR-Match a powerful tool for diverse video editing tasks.

\paragraph{Comparison to other editing methods}

\cref{fig:qualitative} compares STR-Match with recent video editing baselines, showing that our method achieves sharper visual fidelity, tighter foreground–background texture alignment, and more faithful shape transformations.
In the `baby $\rightarrow$ sleeping baby' case, DMT, UniEdit, and VideoGrain tint the infant while leaving the background gray, whereas STR-Match maintains consistent tonality across the entire frame by capturing spatiotemporal pixel relevance through the STR score. 
In the `lotus $\rightarrow$ daisy' example, several baselines either fail to replace the lotus at all or succeed only by unintentionally changing the background.
On the other hand, STR-Match successfully replaces the lotus with high fidelity while preserving the background intact.
The same trend holds on more dynamic contents. 
In the `zebra $\rightarrow$ horse' example, most prior methods either fail to capture the horse’s leg motion (\eg, lifting its leg) or degrade appearance quality, while Ground-A-Video further disrupts scene consistency. 
In contrast, STR-Match faithfully reproduces the motion with high visual fidelity.

Furthermore, STR-Match demonstrates strong performance even in extreme video editing scenarios.
In the `cat $\rightarrow$ basketball' example, most existing methods fail to transform the cat into a basketball, while DMT generates a basketball at the cost of undesired background changes.
Similarly, in the `fish $\rightarrow$ sweet potato' case, DMT and FLATTEN partially modify the object but suffer from background distortion or low fidelity, and other methods fail to perform the edit.
In contrast, STR-Match successfully transforms the object with high visual fidelity while preserving the background.
In summary, STR-Match enables high-fidelity, and flexible shape transformation in video editing while preserving spatiotemporal information.

\vspace{-1mm}
\subsection{Quantitative comparison}
\label{subsec:exp_quant}

\input{tables/quant.tex}

We quantitatively evaluate STR-Match against existing training-free video editing methods for four metrics: temporal consistency~(FC), fidelity to the target prompt~(CS), background preservation~(BL), and motion preservation from the source video~(ME).
STR-Match, with and without binary masks, achieves strong performance, as evidenced by its large area in the radar graph shown in \cref{fig:raidar}.
Notably, compared to T2I-based editing methods, STR-Match demonstrates superior frame consistency, indicating that the proposed STR score effectively captures spatiotemporal pixel relevances from the T2V model.

Furthermore, when comparing STR-Match with masks to UniEdit~(red solid and orange lines), both of which utilize SAM-Track, STR-Match outperforms in all evaluated metrics. 
In the comparison between STR-Match without masks and DMT~(red dashed and green lines), the scores reveal that STR-Match more effectively captures key information from the source video, such as background and motion, while maintaining comparable fidelity. 
This suggests that the STR score achieves a goldilocks balance—preserving essential details from the source video while maintaining the flexibility required for high-fidelity editing—unlike methods that either over-preserve, reducing fidelity, or under-preserve, diminishing faithfulness.

\begin{figure}[t]
    \begin{minipage}[t]{0.55\textwidth}
        \vspace*{0mm}
        \input{tables/ablation1_qual.tex}
    \end{minipage}
    \hspace{3mm}
    \begin{minipage}[t]{0.40\textwidth}
        \vspace*{0mm}
        \centering
        \input{tables/ablation1_quant.tex}
        \vspace{3mm}
        \input{tables/ablation2.tex}
    \end{minipage}
    \vspace{-5mm}
\end{figure}

\subsection{Ablation Study}
\label{subsec:exp_ablation}

\paragraph{Flexibility of STR score}
To evaluate the effectiveness of our proposed STR score, we compare STR-Match with a baseline that optimizes the concatenation of self- and temporal-attention maps.
For the baseline, we adjust the guidance strength $\lambda$ to ensure that the edited video retains key attributes of the source, such as motion dynamics.
\cref{fig:ablation} shows that STR-Match produces significantly higher quality videos compared to the baseline.
For instance, in the `dog $\rightarrow$ cat' case, the baseline method generates oversaturated colored video and in the `turtle $\rightarrow$ shark' case, it fails to alter the sharks' shape into that of turtles.
These two examples illustrate that naïvely using self- and temporal-attention maps as guidance imposes overly strict constraints, whereas the proposed STR score effectively captures key features while providing sufficient flexibility for editing, as it optimizes values that are conceptually derived from the element-wise multiplication of self- and temporal-attention maps.
Moreover, \cref{tab:ablation1} supports this conclusion, as fidelity-related metrics (FC and CS) are higher for our method.
Although the baseline better preserves background and motion, it often fails to transform objects, as demonstrated in \cref{fig:ablation}.



\paragraph{Ablation on the hyperparameter $\lambda$}
$\lambda$ is the only hyperparameter in STR-Match, which controls the guidance strength during optimization.
To investigate its effect, we conduct an ablation study with three values of $\lambda$: ${ 0.005, 0.01, 0.015 }$.
As shown in \cref{tab:ablation2}, we empirically observe that smaller values of $\lambda$ yield higher fidelity scores (FC, CS) but struggle to preserve background and motion dynamics, whereas larger values promote preservation at the cost of fidelity.
To balance these objectives, we adopt $\lambda = 0.01$ for all experiments.

%% file: tables/qual.tex
\begin{figure}[!t]
	\centering
	\setlength{\tabcolsep}{0mm} 
	\renewcommand{\arraystretch}{0.7} 
	\hspace{-3mm}
	\scalebox{1}{
		\begin{tabular}{r @{\extracolsep{1mm}} c @{\extracolsep{1mm}}c @{\extracolsep{1mm}} c @{\extracolsep{1mm}} c @{\extracolsep{0.3mm}} c @{\extracolsep{1mm}}c  @{\extracolsep{0.3mm}} c @{\extracolsep{0.3mm}}c}

			\rotatebox{90}{\makebox[1.0cm][c]{\tiny{\shortstack{Source}}}} &
			\includegraphics[width=0.132\linewidth]{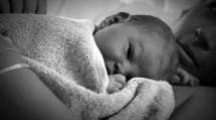} &
			\includegraphics[width=0.132\linewidth]{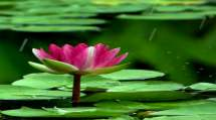} &
			\includegraphics[width=0.132\linewidth]{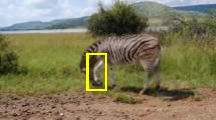} &
			\includegraphics[width=0.132\linewidth]{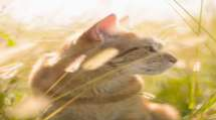} & 
			\includegraphics[width=0.132\linewidth]{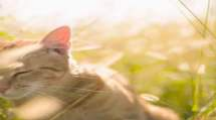} &
			\includegraphics[width=0.132\linewidth]{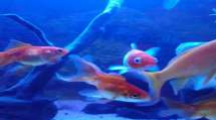} & 
			\includegraphics[width=0.132\linewidth]{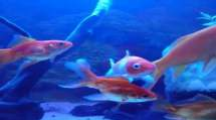} \\
			&
			\multicolumn{1}{c}{\tiny baby $\rightarrow$ sleeping baby} & 
			\multicolumn{1}{c}{\tiny lotus $\rightarrow$ daisy} & 
			\multicolumn{1}{c}{\tiny zebra $\rightarrow$ horse} &
			\multicolumn{2}{c}{\tiny cat $\rightarrow$ basketball} & 
			\multicolumn{2}{c}{\tiny fish $\rightarrow$ sweet potato} \\
			
			\rotatebox{90}{\makebox[1.0cm][c]{\tiny{\shortstack{Ours \\ (w/o Mask)}}}} &
			\includegraphics[width=0.132\linewidth]{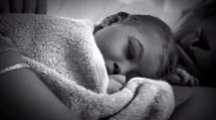} &
			\includegraphics[width=0.132\linewidth]{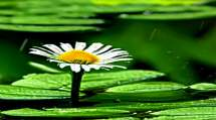} &
			\includegraphics[width=0.132\linewidth]{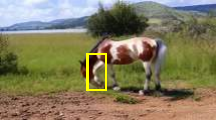} & 
			\includegraphics[width=0.132\linewidth]{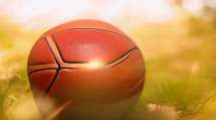} & 
			\includegraphics[width=0.132\linewidth]{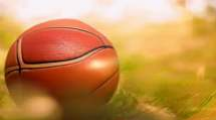} &
			\includegraphics[width=0.132\linewidth]{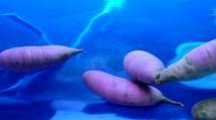} & 
			\includegraphics[width=0.132\linewidth]{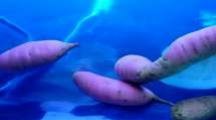}\\

			\rotatebox{90}{\makebox[1.0cm][c]{\tiny{\shortstack{Ours \\ (w/ Mask)}}}} &
			\includegraphics[width=0.132\linewidth]{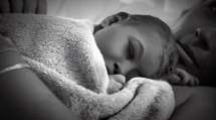} &
			\includegraphics[width=0.132\linewidth]{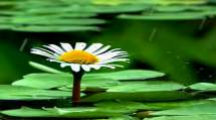} &
			\includegraphics[width=0.132\linewidth]{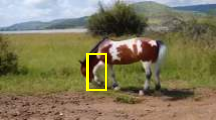} & 
			\includegraphics[width=0.132\linewidth]{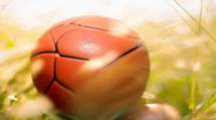} & 
			\includegraphics[width=0.132\linewidth]{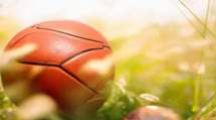} &
			\includegraphics[width=0.132\linewidth]{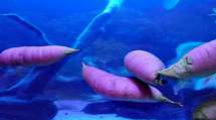} & 
			\includegraphics[width=0.132\linewidth]{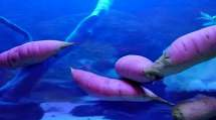}\\

			\vspace{-2mm} \\
			\hdashline
			\vspace{-1mm} \\

			\rotatebox{90}{\makebox[1.0cm][c]{\tiny{\shortstack{DMT}}}} &
			\includegraphics[width=0.132\linewidth]{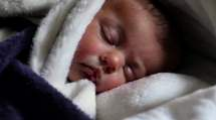} &
			\includegraphics[width=0.132\linewidth]{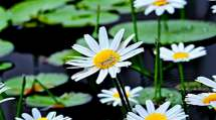} &
			\includegraphics[width=0.132\linewidth]{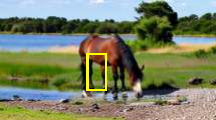} & 
			\includegraphics[width=0.132\linewidth]{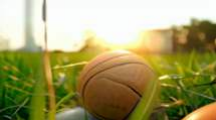} & 
			\includegraphics[width=0.132\linewidth]{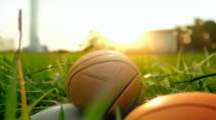} &
			\includegraphics[width=0.132\linewidth]{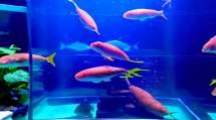} & 
			\includegraphics[width=0.132\linewidth]{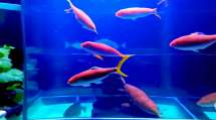}\\

			\rotatebox{90}{\makebox[1.0cm][c]{\tiny{\shortstack{UniEdit}}}} &
			\includegraphics[width=0.132\linewidth]{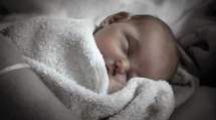} &
			\includegraphics[width=0.132\linewidth]{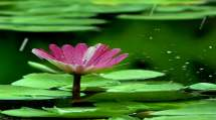} &
			\includegraphics[width=0.132\linewidth]{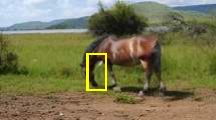} & 
			\includegraphics[width=0.132\linewidth]{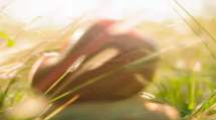} & 
			\includegraphics[width=0.132\linewidth]{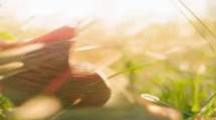} &
			\includegraphics[width=0.13\linewidth]{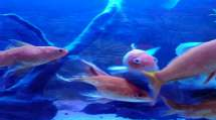} & 
			\includegraphics[width=0.13\linewidth]{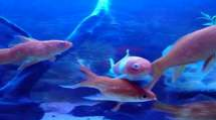}\\

			\rotatebox{90}{\makebox[1.0cm][c]{\tiny{\shortstack{FateZero}}}} &
			\includegraphics[width=0.13\linewidth]{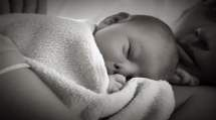} &
			\includegraphics[width=0.13\linewidth]{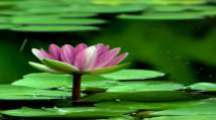} &
			\includegraphics[width=0.13\linewidth]{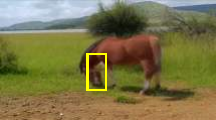} &
			\includegraphics[width=0.13\linewidth]{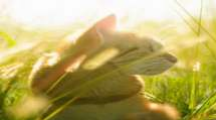} & 
			\includegraphics[width=0.13\linewidth]{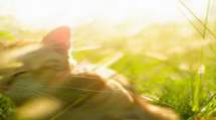} &
			\includegraphics[width=0.13\linewidth]{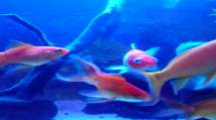} & 
			\includegraphics[width=0.13\linewidth]{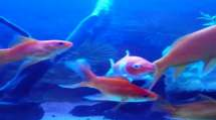}\\

			\rotatebox{90}{\makebox[1.0cm][c]{\tiny{\shortstack{FLATTEN}}}} &
			\includegraphics[width=0.13\linewidth]{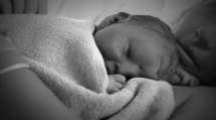} &
			\includegraphics[width=0.13\linewidth]{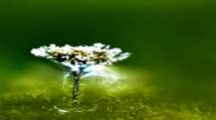} &
			\includegraphics[width=0.13\linewidth]{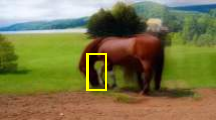} &
			\includegraphics[width=0.13\linewidth]{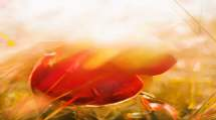} & 
			\includegraphics[width=0.13\linewidth]{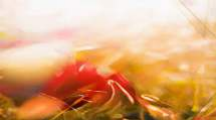} &
			\includegraphics[width=0.13\linewidth]{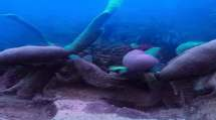} & 
			\includegraphics[width=0.13\linewidth]{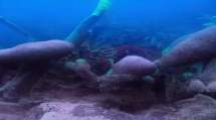} \\

			\rotatebox{90}{\makebox[1.0cm][c]{\tiny{\shortstack{GAV}}}} &
			\includegraphics[width=0.13\linewidth]{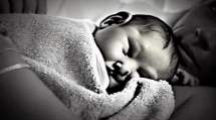} &
			\includegraphics[width=0.13\linewidth]{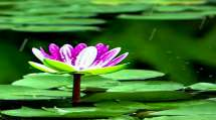} &
			\includegraphics[width=0.13\linewidth]{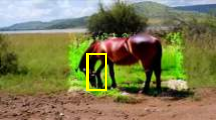} &
			\includegraphics[width=0.13\linewidth]{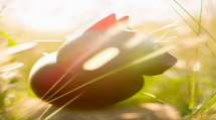} & 
			\includegraphics[width=0.13\linewidth]{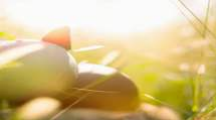} &
			\includegraphics[width=0.13\linewidth]{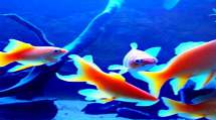} & 
			\includegraphics[width=0.13\linewidth]{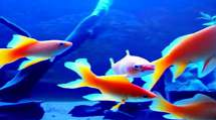} \\

			\rotatebox{90}{\makebox[1.0cm][c]{\tiny{\shortstack{VideoGrain}}}} &
			\includegraphics[width=0.13\linewidth]{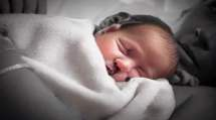} &
			\includegraphics[width=0.13\linewidth]{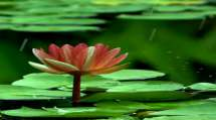} &
			\includegraphics[width=0.13\linewidth]{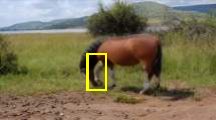} & 
			\includegraphics[width=0.13\linewidth]{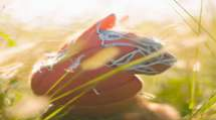} & 
			\includegraphics[width=0.13\linewidth]{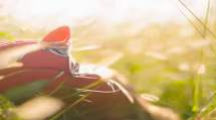} &
			\includegraphics[width=0.13\linewidth]{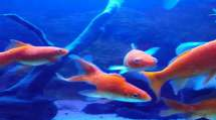} & 
			\includegraphics[width=0.13\linewidth]{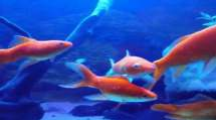} \\
		\end{tabular}
	}
	\caption{
		\textbf{Qualitative comparisons between STR-Match and existing methods.} 
		In each example, STR-Match demonstrates stronger foreground–background texture alignment, higher visual fidelity, better motion alignment, and more flexible shape transformation compared to recent existing methods. 
		Please check our project page for edited videos.
		}
	\label{fig:qualitative}
	\vspace{-2mm}
\end{figure}

%% file: tables/quant.tex

\begin{wrapfigure}{r}{0.45\textwidth}
	\centering
	\includegraphics[width=\linewidth]{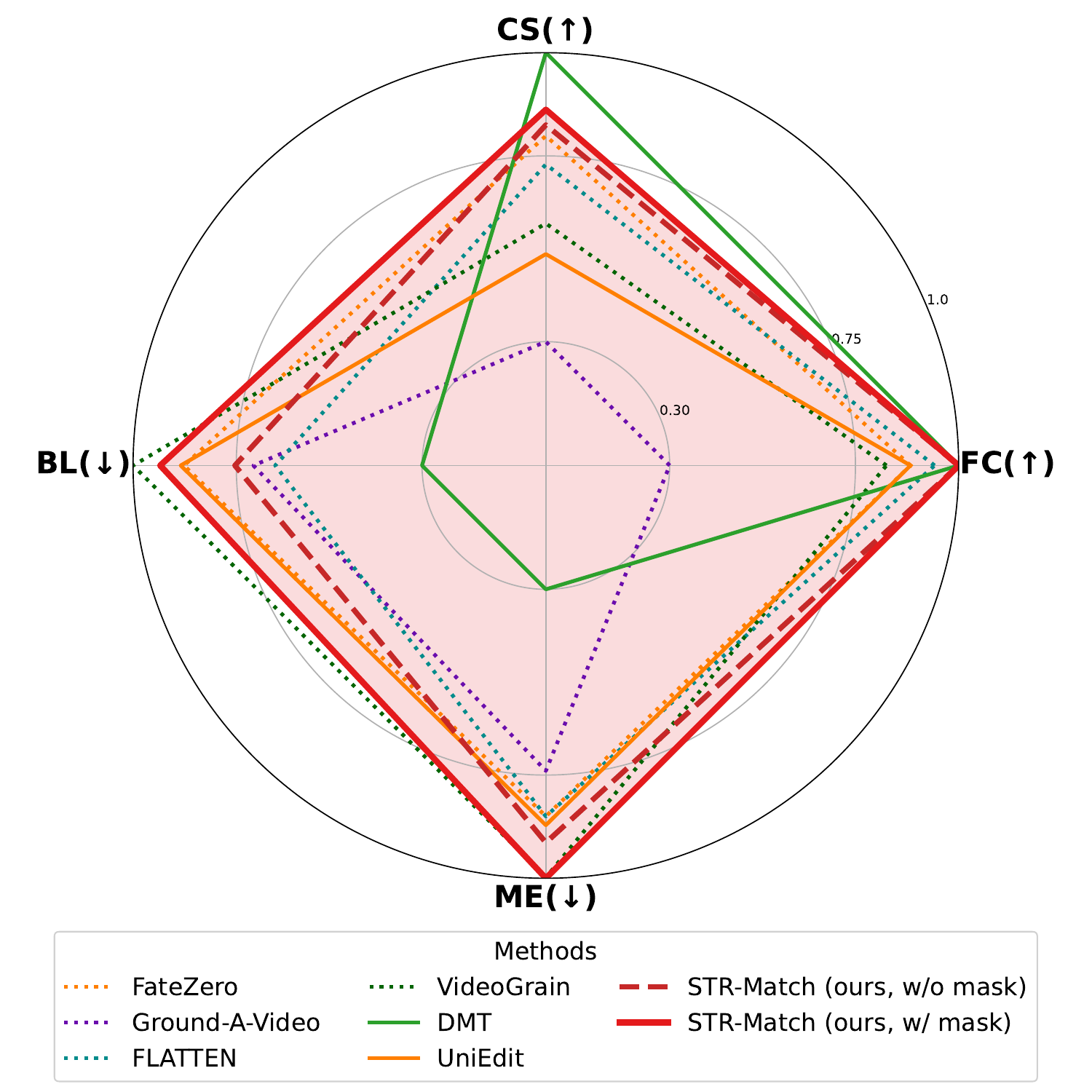}
	\caption{
		Quantitative comparison between STR-Match and existing methods.
		The solid red line is STR-Match with the binary mask, and the dashed red line is STR-Match without binary mask. 
		The solid lines are T2V-based editing methods, while dotted lines are T2I-based methods.
		We provide exact metric numbers and analysis in \cref{tab:details} of \cref{app:quant}.
	}
	\label{fig:raidar}
	\vspace{-5mm}
\end{wrapfigure}

%% file: tables/ablation1_qual.tex

\centering
\setlength{\tabcolsep}{0mm}
\renewcommand{\arraystretch}{0}
\scalebox{0.99}{
	\begin{tabular}{c @{\extracolsep{0.5mm}} c @{\hspace{1mm}}@{\extracolsep{0.5mm}}c @{\hspace{1mm}}@{\extracolsep{0.5mm}}c}

		\rotatebox{90}{\hspace{6mm} \scriptsize{\shortstack{Source}}} &
		\includegraphics[width=0.45\textwidth]{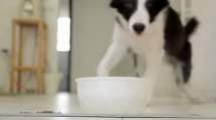} &
		\includegraphics[width=0.45\textwidth]{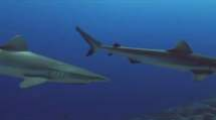} \\
		\vspace{2mm} \\
		
		& \multicolumn{1}{c}{\footnotesize dog $\rightarrow$ cat} & 
		\multicolumn{1}{c}{\footnotesize shark $\rightarrow$ turtle} \\
		\vspace{2mm}\\
		
		\rotatebox{90}{\hspace{4mm} \scriptsize{\shortstack{Baseline \\ (w/o Mask)}}} &
		\includegraphics[width=0.45\textwidth]{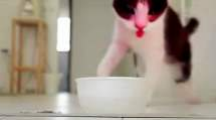} &
		\includegraphics[width=0.45\textwidth]{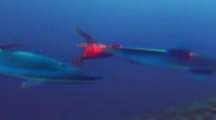} \\
		\vspace{2mm} \\
		
		\rotatebox{90}{\hspace{3.5mm} \scriptsize{\shortstack{Ours \\ (w/o Mask)}}} &
		\includegraphics[width=0.45\textwidth]{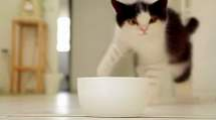} &
		\includegraphics[width=0.45\textwidth]{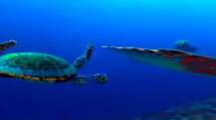} \\
		\vspace{2mm} \\
		
	\end{tabular}
}
\captionof{figure}{Quantitative comparision between STR-Match and the baseline.} 
\label{fig:ablation}

%% file: tables/ablation1_quant.tex

\renewcommand{\arraystretch}{1.3}
\setlength{\tabcolsep}{1.5mm}

\captionof{table}{
	Quantitative comparison between STR-Match and the baseline without mask.
	Bold numbers indicate the better score for each metric.
}
\label{tab:ablation1}

\scalebox{0.85}{
\centering
\begin{tabular}{lcccc}
	\toprule
	Method & FC~($\uparrow$) & CS~($\uparrow$) & BL~($\downarrow$) & ME~($\downarrow$) \\
	\hline
	Baseline & 0.979 & 31.24 & \black{0.117} & \black{2.293} \\
	Ours & \black{0.981} & \black{31.61} & 0.216 & 2.402 \\
	\bottomrule
\end{tabular}
}

%% file: tables/ablation2.tex

\renewcommand{\arraystretch}{1.3}
\setlength{\tabcolsep}{1.5mm}

\captionof{table}{
	Ablation study on $\lambda$ values. 
	Bold black and red numbers indicate the best and second-best scores for each metric, respectively.
}
\label{tab:ablation2}
\vspace{2mm}

\scalebox{0.85}{
\centering
\begin{tabular}{lcccc}
	\toprule
	\hspace{2mm} $\lambda$ & FC~($\uparrow$) & CS~($\uparrow$) & BL~($\downarrow$) & ME~($\downarrow$) \\
	\hline
	$0.005$ & \black{0.982} & \bred{31.60} & 0.271 & 3.120 \\
	$0.01$ & \bred{0.981} & \black{31.61} & \bred{0.216} & \bred{2.402} \\
	$0.015$ & 0.979 & 31.33 & \black{0.196} & \black{2.225} \\
	\bottomrule
\end{tabular}
}

%% file: sections/conclusion.tex

\vspace{0.5mm}
\section{Conclusion}

In this work, we propose a novel spatiotemporal modeling approach that relates to key limitations in existing video editing methods—such as frame inconsistency, motion distortion, visual artifacts, and notably, limited performance in challenging settings like large-gap domain shifts.
To overcome these challenges, we propose the STR score, a spatiotemporal pixel relevance score that captures essential video attributes.
Notably, it is computed solely from the self- and temporal-attention maps of a pretrained text-to-video (T2V) diffusion model, requiring no additional training or external models.
By integrating the STR score into a latent optimization framework alongside a latent mask strategy, we introduce STR-Match, a zero-shot, training-free video editing algorithm that is compatible with any T2V model incorporating temporal modules.
Extensive experiments show that STR-Match consistently outperforms existing training-free methods across all quantitative metrics.
Moreover, it generates videos with substantially improved visual quality, supporting realistic and flexible domain transformatiomn, preserved motion dynamics, and strong temporal consistency.
These results demonstrate both the effectiveness and generalizability of STR-Match, establishing it as a new state-of-the-art baseline for training-free text-guided video editing.

%% file: sections/supple_.tex

\section*{\Large{Appendix}}

\section{Qualitative results}
\label{app:qual}

\subsection{Additional comparisons with other methods}

We provide video files on our project page that showcase a variety of video editing examples—ranging from typical cases with minimal domain shifts to more challenging ones with significant shape transformations, as illustrated in \cref{fig:supple_qual1,fig:supple_qual2}.
For instances involving extreme shape transformations~(\eg `cat $\rightarrow$ dragon', `goldfish $\rightarrow$ snake'), many competing methods distort the target objects to conform to the shape of the source, resulting in unnatural edited outputs. 
In the cases of extreme domain change~(\eg `cat $\rightarrow$ robot dog', `goldfish $\rightarrow$ donuts'), few other methods transfer only a part of the intended concept while the majority fail to perform any meaningful editing.
In contrast, STR-Match consistently delivers successful video edits across these challenging scenarios, highlighting its flexibility and robust editing capabilities.
We strongly encourage readers to view the HTML file included in the zip archive for a more comprehensive understanding of STR-Match’s editing capabilities.

\subsection{STR-Match with Zeroscope}

\input{tables/supple_zeroscope}

Our proposed algorithm leverages the pretrained T2V model equipped with temporal modules.
While we utilize LaVie~\cite{Lavie} as pretrained T2V model for the most of the experiment, STR-Match can also be applied to other T2V models, such as Zeroscope~\footnote{\url{https://huggingface.co/cerspense/zeroscope_v2_576w}}.
\cref{fig:supple_zeroscope} illustrates the results of STR-Match using Zeroscope as the base model.
The results demonstrate that STR-Match can effectively edit videos with Zeroscope, achieving similar performance to LaVie.

\section{Quantitative metrics and model dependencies}
\label{app:quant}

\cref{tab:details} presents evaluation metrics used in the radar graph shown in Figure 5 of Section 5.3 in the main paper along with the base diffusion models and external models used by each method.
Overall, the proposed method, STR-Match, whether applied with and without masks, achieves a balanced and superior performance across all metrics compared to other methods. 
Notably, while DMT generates high quality videos~(evidenced by strong FC and CS), these outputs often lack fidelity to the source video (reflected in poor scores for BL and ME).
Although we have provided the quantitative metrics, we encourage readers to consult the qualitatve results in Figure 4 in the main paper, \cref{app:qual}, and supplementary material, as these metrics are incomplete and often fail to reflect the true quality of videos. 

\input{tables/supple_details}

\section{Limitations}
\label{app:limitation}


While STR-Match produces satisfying editing results, even in the challenging scenarios like flexible shape transformations, it still has some limitations.
One limitation is its inability to edit multiple objects into different targets simultaneously. 
Although a workaround exists—editing each object individually with its corresponding mask—this approach is highly inefficient. 
Additionally, while the method supports flexible shape transformations, it produces suboptimal results when the object’s size varies significantly. 
We plan to address remaining limitations in future work.


\section{Societal Impact}
\label{app:societal}

STR-Match is a training-free video editing algorithm that leverages pretrained T2V models. 
Since it relies heavily on these pretrained models, there is a potential risk of generating videos with unintended or inappropriate contents. 
However, we believe this issue can be indirectly mitigated by carefully controlling the training data used for the underlying T2V models.

\input{tables/supple_qual1}
\input{tables/supple_qual2}

%% file: tables/supple_zeroscope.tex
\begin{figure}[h]
	\centering
	\setlength{\tabcolsep}{0mm} 
	\renewcommand{\arraystretch}{0.5} 
	\scalebox{1.04}{
		\begin{tabular}{c @{\extracolsep{0.5mm}}c @{\extracolsep{1.5mm}} c @{\extracolsep{0.5mm}} c @{\extracolsep{1.5mm}} c @{\extracolsep{0.5mm}}c}

			\includegraphics[width=0.15\linewidth]{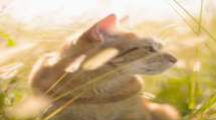} &
			\includegraphics[width=0.15\linewidth]{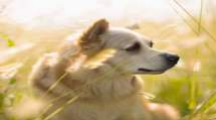} &
			\includegraphics[width=0.15\linewidth]{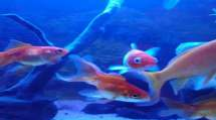} &
			\includegraphics[width=0.15\linewidth]{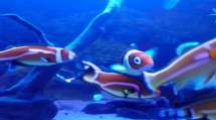} &
			\includegraphics[width=0.15\linewidth]{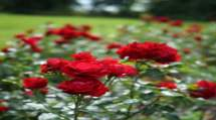} &
			\includegraphics[width=0.15\linewidth]{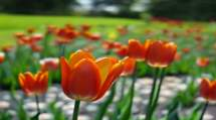} \\
			
			\multicolumn{2}{c}{\tiny cat $\rightarrow$ dog} & 
			\multicolumn{2}{c}{\tiny goldfish $\rightarrow$ clownfish} & 
			\multicolumn{2}{c}{\tiny red roses $\rightarrow$ orange tulips} \\
		\end{tabular}
	}
	\caption{
		\textbf{Qualitative results of STR-Match using Zeroscope.} 
		STR-Match can be applied to Zeroscope, achieving similar performance to LaVie.
		}
	\label{fig:supple_zeroscope}  
\end{figure}

%% file: tables/supple_details.tex

\begin{table}[h]
	\centering
	\caption{
		\textbf{Quantitative comparison and model dependencies between STR-Match and existing methods.}
		For quantitative metrics, bold black and red numbers indicate the best and second-best performance for each metric, respectively.
		Note that FC~(Frame Consistency) and CS~(CLIP Similarity) are higher-is-better metrics, while BL~(BG-LPIPS) and ME~(Motion Error) are lower-is-better.
	}
	\label{tab:details}
	\setlength\tabcolsep{4pt} 
	\vspace{2mm}
	\renewcommand{\arraystretch}{1.5}
	\scalebox{0.8}{
			\begin{tabular}{lllcccc}
				\toprule
				Method & Base model & External model & FC~($\uparrow$) & CS~($\uparrow$) & BL~($\downarrow$) & ME~($\downarrow$) \\ 
				\hline
				
				{FateZero~\cite{fatezero}} & T2I~(sd1.4~\tablefootnote{\url{https://huggingface.co/CompVis/stable-diffusion-v1-4}}) & 
				-- 
				&0.979 &31.56 & 0.139 &2.749 \\
				
				{FLATTEN~\cite{flatten}} & T2I~(sd2.1~\tablefootnote{\url{https://huggingface.co/stabilityai/stable-diffusion-2-1}}) & 
				RAFT 
				&\bred{0.980} &31.43 &0.277 &2.748 \\

				{VideoGrain~\cite{videograin}} & T2I~(sd1.5~\tablefootnote{\url{https://huggingface.co/stable-diffusion-v1-5/stable-diffusion-v1-5}}) & 
				RAFT, SAM-Track, ControlNet 
				&0.978 &31.16 &\black{0.062} &\bred{1.943} \\
				
				{Ground-A-Video~\cite{gav}} & T2I~(sd1.5) & 
				\parbox[c]{5cm}{ControlNet~\cite{controlnet}, GLIGEN~\cite{gligen}\\ RAFT~\cite{raft}, ZoeDepth~\cite{zoedepth}, OWL-ViT~\cite{bbox}} 
				&0.969 & 30.62 &0.244 &3.348 \\
				\hline
				
				{DMT~\cite{dmt}} & T2V~(LaVie) & -- 
				&\black{0.981} &\black{31.94} &0.499 &5.741 \\
				
				{UniEdit~\cite{uniedit}} & T2V~(LaVie) & SAM-Track 
				& 0.979 & 31.02 & 0.134 & 2.632 \\ 
	
				{STR-Match (Ours, w/o mask)} & T2V~(LaVie) & -- 
				&\black{0.981} & 31.61 & 0.216 & 2.402 \\
				
				{STR-Match (Ours, w/ mask)} & T2V~(LaVie) & SAM-Track 
				&\black{0.981} &\bred{31.68} &\bred{0.103} & \black{1.932}\\
				
				\bottomrule 
			\end{tabular}
			}
\end{table}

				
				

				
				
				
	
				
				

%% file: tables/supple_qual1.tex
\begin{figure}[b]
	\centering
	\setlength{\tabcolsep}{0mm} 
	\renewcommand{\arraystretch}{0.5} 
	\scalebox{1}{
		\begin{tabular}{r @{\extracolsep{1.5mm}} c @{\extracolsep{0.5mm}}c @{\extracolsep{1.5mm}} c @{\extracolsep{0.5mm}} c @{\extracolsep{1.5mm}} c @{\extracolsep{0.5mm}}c}

			\rotatebox{90}{\makebox[1.2cm][c]{\tiny{\shortstack{Source}}}} &
			\includegraphics[width=0.15\linewidth]{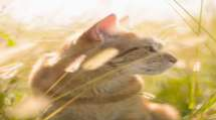} &
			\includegraphics[width=0.15\linewidth]{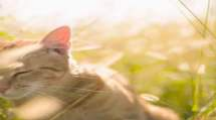} &
			\includegraphics[width=0.15\linewidth]{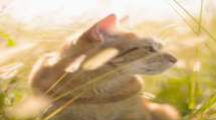} &
			\includegraphics[width=0.15\linewidth]{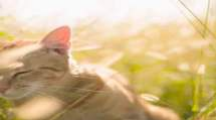} &
			\includegraphics[width=0.15\linewidth]{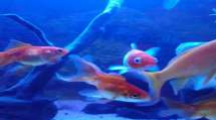} &
			\includegraphics[width=0.15\linewidth]{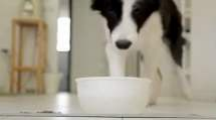} \\
			 
			& 
			\multicolumn{2}{c}{\tiny cat $\rightarrow$ dragon} & 
			\multicolumn{2}{c}{\tiny cat $\rightarrow$ robot dog} & 
			\multicolumn{1}{c}{\tiny goldfish $\rightarrow$ snake} & 
			\multicolumn{1}{c}{\tiny dog $\rightarrow$ cat} \\
			
			\rotatebox{90}{\makebox[1.2cm][c]{\tiny{\shortstack{Ours \\ (w/o Mask)}}}} &
			\includegraphics[width=0.15\linewidth]{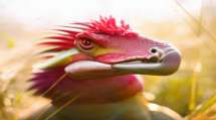} &
			\includegraphics[width=0.15\linewidth]{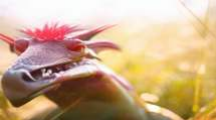} &
			\includegraphics[width=0.15\linewidth]{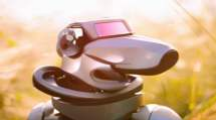} &
			\includegraphics[width=0.15\linewidth]{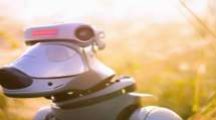} &
			\includegraphics[width=0.15\linewidth]{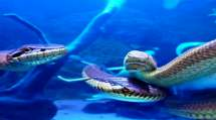} & 
			\includegraphics[width=0.15\linewidth]{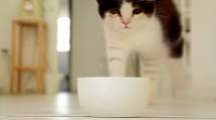} \\

			\rotatebox{90}{\makebox[1.2cm][c]{\tiny{\shortstack{Ours \\ (w/ Mask)}}}} &
			\includegraphics[width=0.15\linewidth]{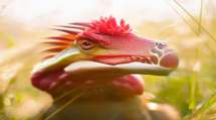} &
			\includegraphics[width=0.15\linewidth]{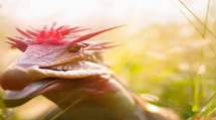} &
			\includegraphics[width=0.15\linewidth]{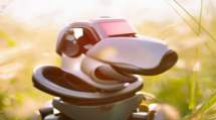} &
			\includegraphics[width=0.15\linewidth]{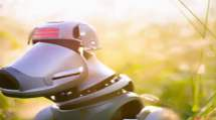} &
			\includegraphics[width=0.15\linewidth]{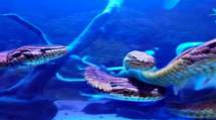} & 
			\includegraphics[width=0.15\linewidth]{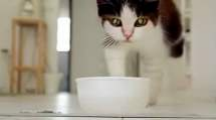} \\

			\vspace{0.5mm} \\
			\hdashline
			\vspace{0.5mm} \\

			\rotatebox{90}{\makebox[1.2cm][c]{\tiny{\shortstack{DMT}}}} &
			\includegraphics[width=0.15\linewidth]{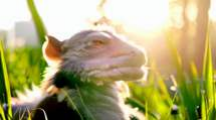} &
			\includegraphics[width=0.15\linewidth]{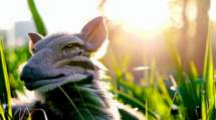} &
			\includegraphics[width=0.15\linewidth]{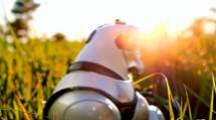} &
			\includegraphics[width=0.15\linewidth]{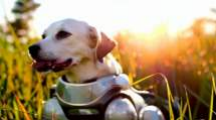} &
			\includegraphics[width=0.15\linewidth]{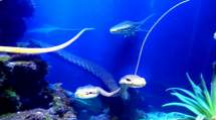} & 
			\includegraphics[width=0.15\linewidth]{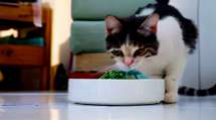} \\

			\rotatebox{90}{\makebox[1.2cm][c]{\tiny{\shortstack{UniEdit}}}} &
			\includegraphics[width=0.15\linewidth]{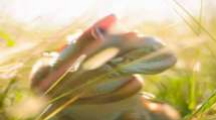} &
			\includegraphics[width=0.15\linewidth]{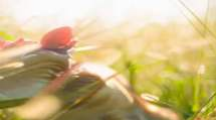} &
			\includegraphics[width=0.15\linewidth]{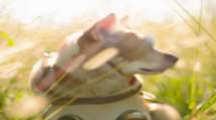} &
			\includegraphics[width=0.15\linewidth]{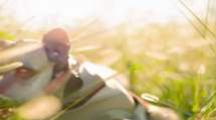} &
			\includegraphics[width=0.15\linewidth]{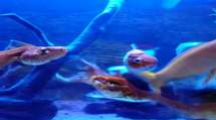} &
			\includegraphics[width=0.15\linewidth]{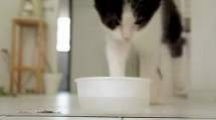}  \\

			\rotatebox{90}{\makebox[1.2cm][c]{\tiny{\shortstack{FateZero}}}} &
			\includegraphics[width=0.15\linewidth]{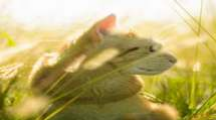} &
			\includegraphics[width=0.15\linewidth]{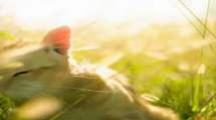} &
			\includegraphics[width=0.15\linewidth]{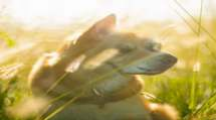} &
			\includegraphics[width=0.15\linewidth]{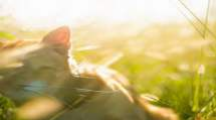} &
			\includegraphics[width=0.15\linewidth]{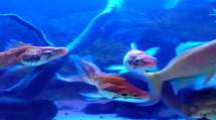} &
			\includegraphics[width=0.15\linewidth]{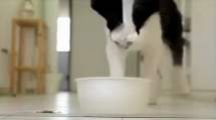} \\

			\rotatebox{90}{\makebox[1.2cm][c]{\tiny{\shortstack{FLATTEN}}}} &
			\includegraphics[width=0.15\linewidth]{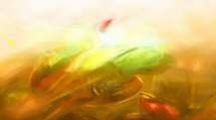} &
			\includegraphics[width=0.15\linewidth]{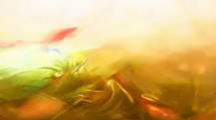} &
			\includegraphics[width=0.15\linewidth]{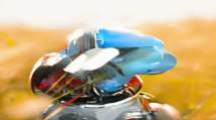} &
			\includegraphics[width=0.15\linewidth]{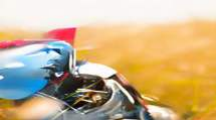} &
			\includegraphics[width=0.15\linewidth]{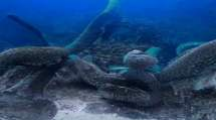} &
			\includegraphics[width=0.15\linewidth]{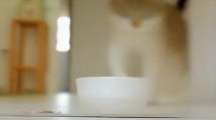}  \\

			\rotatebox{90}{\makebox[1.2cm][c]{\tiny{\shortstack{GAV}}}} &
			\includegraphics[width=0.15\linewidth]{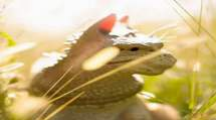} &
			\includegraphics[width=0.15\linewidth]{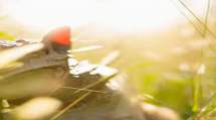} &
			\includegraphics[width=0.15\linewidth]{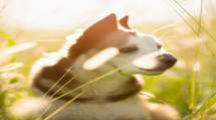} &
			\includegraphics[width=0.15\linewidth]{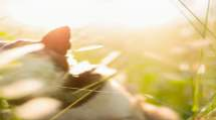} &
			\includegraphics[width=0.15\linewidth]{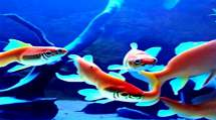} &
			\includegraphics[width=0.15\linewidth]{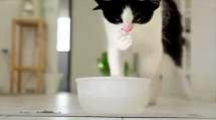} \\

			\rotatebox{90}{\makebox[1.2cm][c]{\tiny{\shortstack{VideoGrain}}}} &
			\includegraphics[width=0.15\linewidth]{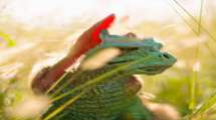} &
			\includegraphics[width=0.15\linewidth]{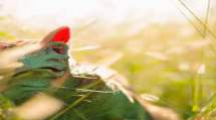} &
			\includegraphics[width=0.15\linewidth]{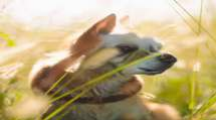} &
			\includegraphics[width=0.15\linewidth]{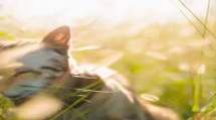} &
			\includegraphics[width=0.15\linewidth]{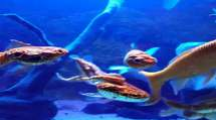} &
			\includegraphics[width=0.15\linewidth]{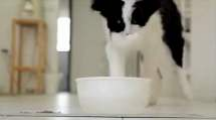} \\
		\end{tabular}
	}
	\caption{
		\textbf{Additional qualitative comparisons between STR-Match and existing methods.} 
		This figure illustrates the performance of STR-Match in challenging scenarios, including cat $\rightarrow$ dragon, cat $\rightarrow$ robot dog, goldfish $\rightarrow$ snake, and dog $\rightarrow$ cat.
		}
	\label{fig:supple_qual1}  
\end{figure}

%% file: tables/supple_qual2.tex
\begin{figure}[b]
	\centering
	\setlength{\tabcolsep}{0mm} 
	\renewcommand{\arraystretch}{0.5} 
	\scalebox{1}{
		\begin{tabular}{r @{\extracolsep{1.5mm}} c @{\extracolsep{0.5mm}}c @{\extracolsep{1.5mm}} c @{\extracolsep{0.5mm}} c @{\extracolsep{1.5mm}} c @{\extracolsep{0.5mm}}c}

			\rotatebox{90}{\makebox[1.2cm][c]{\tiny{\shortstack{Source}}}} &
			\includegraphics[width=0.15\linewidth]{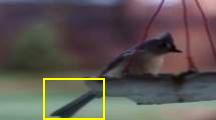} &
			\includegraphics[width=0.15\linewidth]{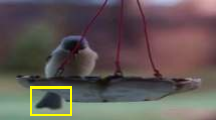} &
			\includegraphics[width=0.15\linewidth]{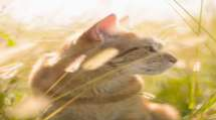} &
			\includegraphics[width=0.15\linewidth]{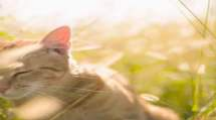} &
			\includegraphics[width=0.15\linewidth]{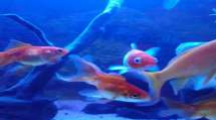} &
			\includegraphics[width=0.15\linewidth]{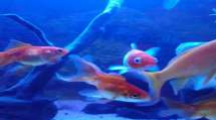} \\
			 
			& 
			\multicolumn{2}{c}{\tiny bird $\rightarrow$ cat} & 
			\multicolumn{2}{c}{\tiny cat $\rightarrow$ giraffe} & 
			\multicolumn{1}{c}{\tiny goldfish $\rightarrow$ donuts} & 
			\multicolumn{1}{c}{\tiny goldfish $\rightarrow$ clownfish} \\
			
			\rotatebox{90}{\makebox[1.2cm][c]{\tiny{\shortstack{Ours \\ (w/o Mask)}}}} &
			\includegraphics[width=0.15\linewidth]{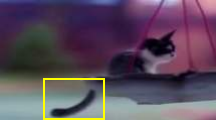} &
			\includegraphics[width=0.15\linewidth]{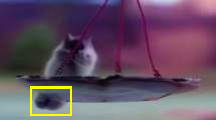} &
			\includegraphics[width=0.15\linewidth]{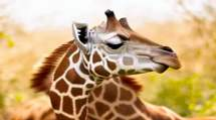} &
			\includegraphics[width=0.15\linewidth]{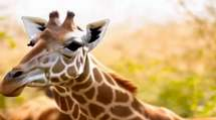} &
			\includegraphics[width=0.15\linewidth]{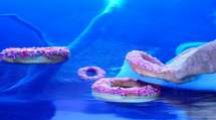} & 
			\includegraphics[width=0.15\linewidth]{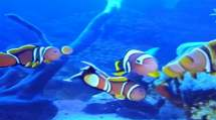} \\

			\rotatebox{90}{\makebox[1.2cm][c]{\tiny{\shortstack{Ours \\ (w/ Mask)}}}} &
			\includegraphics[width=0.15\linewidth]{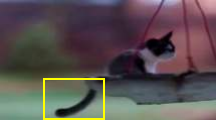} &
			\includegraphics[width=0.15\linewidth]{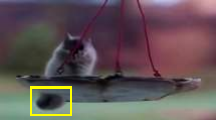} &
			\includegraphics[width=0.15\linewidth]{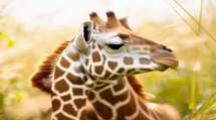} &
			\includegraphics[width=0.15\linewidth]{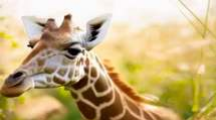} &
			\includegraphics[width=0.15\linewidth]{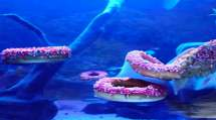} & 
			\includegraphics[width=0.15\linewidth]{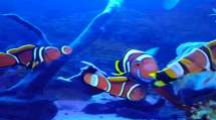} \\

			\vspace{0.5mm} \\
			\hdashline
			\vspace{0.5mm} \\

			\rotatebox{90}{\makebox[1.2cm][c]{\tiny{\shortstack{DMT}}}} &
			\includegraphics[width=0.15\linewidth]{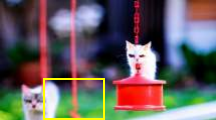} &
			\includegraphics[width=0.15\linewidth]{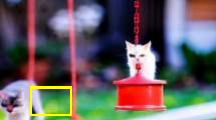} &
			\includegraphics[width=0.15\linewidth]{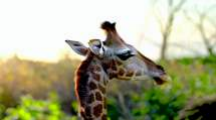} &
			\includegraphics[width=0.15\linewidth]{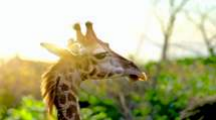} &
			\includegraphics[width=0.15\linewidth]{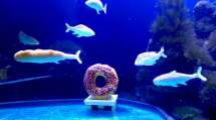} & 
			\includegraphics[width=0.15\linewidth]{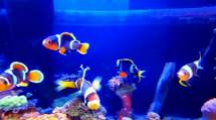} \\

			\rotatebox{90}{\makebox[1.2cm][c]{\tiny{\shortstack{UniEdit}}}} &
			\includegraphics[width=0.15\linewidth]{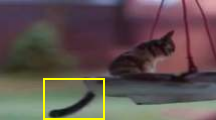} &
			\includegraphics[width=0.15\linewidth]{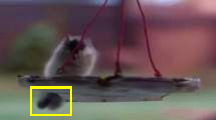} &
			\includegraphics[width=0.15\linewidth]{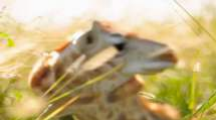} &
			\includegraphics[width=0.15\linewidth]{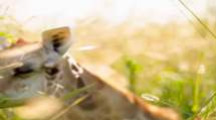} &
			\includegraphics[width=0.15\linewidth]{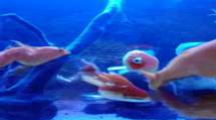} &
			\includegraphics[width=0.15\linewidth]{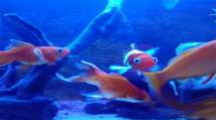}  \\

			\rotatebox{90}{\makebox[1.2cm][c]{\tiny{\shortstack{FateZero}}}} &
			\includegraphics[width=0.15\linewidth]{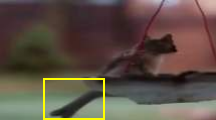} &
			\includegraphics[width=0.15\linewidth]{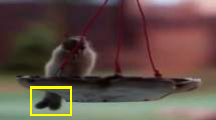} &
			\includegraphics[width=0.15\linewidth]{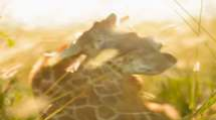} &
			\includegraphics[width=0.15\linewidth]{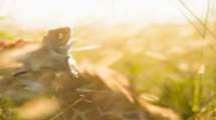} &
			\includegraphics[width=0.15\linewidth]{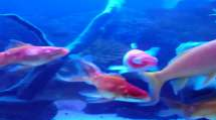} &
			\includegraphics[width=0.15\linewidth]{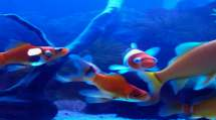} \\

			\rotatebox{90}{\makebox[1.2cm][c]{\tiny{\shortstack{FLATTEN}}}} &
			\includegraphics[width=0.15\linewidth]{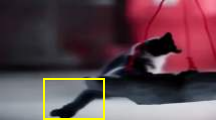} &
			\includegraphics[width=0.15\linewidth]{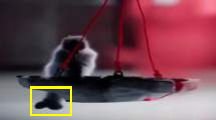} &
			\includegraphics[width=0.15\linewidth]{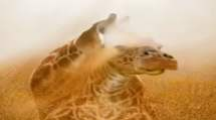} &
			\includegraphics[width=0.15\linewidth]{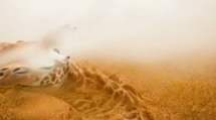} &
			\includegraphics[width=0.15\linewidth]{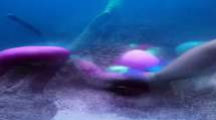} &
			\includegraphics[width=0.15\linewidth]{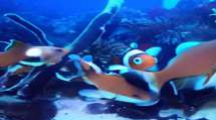}  \\

			\rotatebox{90}{\makebox[1.2cm][c]{\tiny{\shortstack{GAV}}}} &
			\includegraphics[width=0.15\linewidth]{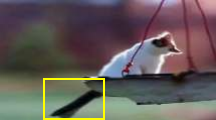} &
			\includegraphics[width=0.15\linewidth]{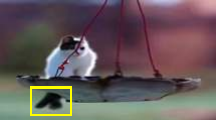} &
			\includegraphics[width=0.15\linewidth]{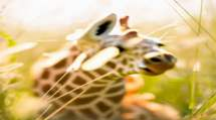} &
			\includegraphics[width=0.15\linewidth]{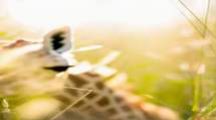} &
			\includegraphics[width=0.15\linewidth]{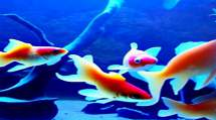} &
			\includegraphics[width=0.15\linewidth]{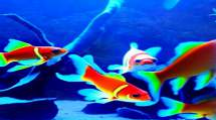} \\

			\rotatebox{90}{\makebox[1.2cm][c]{\tiny{\shortstack{VideoGrain}}}} &
			\includegraphics[width=0.15\linewidth]{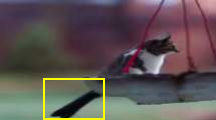} &
			\includegraphics[width=0.15\linewidth]{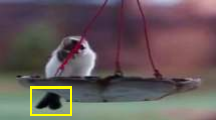} &
			\includegraphics[width=0.15\linewidth]{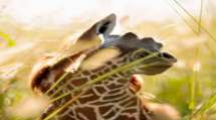} &
			\includegraphics[width=0.15\linewidth]{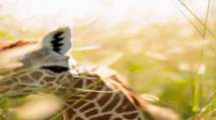} &
			\includegraphics[width=0.15\linewidth]{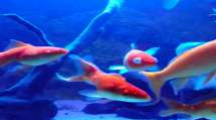} &
			\includegraphics[width=0.15\linewidth]{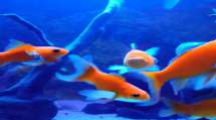} \\
		\end{tabular}
	}
	\caption{
		\textbf{Qualitative comparisons between STR-Match and existing methods.} 
		This figure illustrates the performance of STR-Match in challenging scenarios, including bird $\rightarrow$ cat, cat $\rightarrow$ giraffe, goldfish $\rightarrow$ donuts, and goldfish $\rightarrow$ clownfish.
		}
	\label{fig:supple_qual2}  
\end{figure}